\documentclass{llncs}
\usepackage{wrapfig}
\usepackage{graphicx}
\usepackage{color}
\usepackage[misc]{ifsym}

\begin{document}

\title{Generative Invertible Networks (GIN): Pathophysiology-Interpretable Feature Mapping and Virtual Patient Generation}

\titlerunning{GIN for virtual patients}  
\vspace{-15pt}
\author{Jialei CHEN\inst{1} \inst{2} \textsuperscript{(\Letter)} \and Yujia XIE\inst{3}  \and Kan WANG\inst{1} \inst{2} \and Zih Huei WANG \inst{4}  \and Geet LAHOTI\inst{1} \inst{2} \and Chuck ZHANG\inst{1} \inst{2} \and Mani A VANNAN\inst{5} \and Ben WANG\inst{1} \inst{2} \inst{6}  \and Zhen QIAN\inst{5} \textsuperscript{(\Letter)}}
\authorrunning{Jialei CHEN et al.} 
%
\tocauthor{Jialei CHEN, Yujia XIE, Zhen QIAN, Kan WANG, Mani VANNAN, Chuck ZHANG, Ben WANG}
\vspace{-15pt}

\institute{Georgia Tech Manufacturing Institute, Georgia Institute of Technology,\\ \email{jialei.chen@gatech.edu}  \\ \and H. Milton Stewart School of Industrial and Systems Engineering, Georgia Tech,\and School of Computational Science and Engineering, Georgia Tech, \\ \and  Department of Industrial Engineering and Engineering Management, \\ National Tsing Hua University, \and Marcus Heart Valve Center, Piedmont Heart Institute,  \\ \email{Zhen.Qian@piedmont.org}  \\ \and  School of Materials Science and Engineering, Georgia Tech.\\ }

\maketitle              
\vspace{-12pt}
\begin{abstract}
\vspace{-12pt}
Machine learning methods play increasingly important roles in pre-procedural planning for complex surgeries and interventions.  
Very often, however, researchers find the historical records of emerging surgical techniques, such as the transcatheter aortic valve replacement (TAVR), are highly scarce in quantity.
In this paper, we address this challenge by proposing novel generative invertible networks (GIN) to select features and generate high-quality \textit{virtual patients} that may potentially serve as an additional data source for machine learning.
Combining a convolutional neural network (CNN) and generative adversarial networks (GAN), GIN discovers the pathophysiologic meaning of the feature space.
Moreover, a test of predicting the surgical outcome directly using the selected features results in a high accuracy of 81.55\%, which suggests little pathophysiologic information has been lost while conducting the feature selection.
This demonstrates GIN can generate virtual patients not only visually authentic but also pathophysiologically interpretable.
\vspace{-8pt}
\keywords{Virtual patients, Generative neural networks}
\end{abstract}
\vspace{-28pt}

\section{Introduction}

\vspace{-8pt}

For pre-surgical planning of complex surgeries and interventions, it remains difficult to build a comprehensive pathophysiology-based model incorporating the dynamic interactions between the human body and the medical device. 
Developing machine learning models from historical surgical data to help predict and optimize the surgical outcome has become a promising alternative. 
In literature, machine learning methods (e.g., random forests \cite{statnikov2008comprehensive}, logistic regression \cite{kim2000permutation}) have been used for various prediction purposes based on pre-selected features, while recently, deep learning methods (e.g., convolutional neural networks \cite{shin2016deep}) have emerged for feature selection and outcome prediction directly based on the input images. However, the key challenge to most surgery-related machine learning problems is that, while existing machine learning methods typically require large amounts of data, the dataset available consists of data from only a limited number of patients, which is usually too small for training considering the high dimensional input data (usually a fusion of medical images and clinical records). 
Furthermore, the highly unbalanced prediction input (e.g., age, blood pressure) and output (e.g., surgical outcome) add another layer of difficulty.  
In short, machine learning methods based on existing surgical records have limitations, and an enhancement of data size is imperative.

One immediate method to enlarge the data size is data augmentation \cite{greenland2001data}, including image translation, rotation, changing in brightness and tune, etc.
Nevertheless, most image augmentation methods used in natural images may impose alterations with pathophysiologic significance to medical images. For example, in CT scans, image intensity corresponds to specific substances of human tissue, alterations of which may change the tissue type and lead to a different surgical outcome. This difference limits the effectiveness of image augmentation in medical images. 
Meanwhile, a bypass method that is also widely adopted is transfer learning technique \cite{goodfellow2016deep}. Researchers try to adapt the pre-trained model from natural images and modify a small amount of the model parameters for medical applications with less training data \cite{shin2016deep}. 
Yet a strong assumption of transfer learning is that the image features learnt from natural images would work similarly in medical images.
For the prediction of surgical outcomes, the rationality of that is not clear, because a surgery involves a complex and dynamic interaction between the human anatomy and the surgical device, and the visual cues extracted from the medical images may not be sufficient for such a prediction. In one of our recent work, the predictive performance for transcatheter aortic valve replacement (TAVR) outcome using transfer learning is inferior to a CNN learnt from scratch \cite{joy}. 
This urges us to explore other possibilities.

Another way of data size enhancement is to generate \textit{virtual patients}. Different from some literature, here it refers to the digital models that mimic the patient organ but are not exactly the same as any real patients \cite{qian2017quantitative}. The virtual patients can be 3D printed for a bench-top surgical simulation to assess surgical outcomes just like in the real patients as an enhancement to the dataset \cite{wang2015designable}.
While medical image simulation based on a 4D extended cardiac-torso (XCAT) phantom is widely investigated \cite{segars20104d}, a complete \textit{generative} model from scratch is lacking in medical literature.
Some models from the machine learning community have the potential for virtual patient generation, including restrict Boltzmann machine (RBM) and variational auto-encoder (VAE) \cite{goodfellow2016deep}. Yet these methods usually lead to sever blurriness in generated images. 
Recently, a deep learning framework, generative adversarial networks (GAN) was proposed to generate high-quality images, based on the distribution of the training images (see Section 2.2), which can be authentic enough to fool human eyes \cite{arjovsky2017wasserstein}. 
A straightforward idea is to adapt GAN for virtual patient generation. However, all of the generative methods above result in generating virtual patients that \textit{visually} look like real patients, but with unclear pathophysiologic meanings.

In this work, we proposed a novel, deep learning framework - generative invertible networks (GIN) to extract the features from the real patients and generate virtual patients, which were both visually and pathophysiologically plausible, using the features
(see Section 2). Specifically, GIN tries to find the feature mapping from the high-dimensional human issue/organ space (represented by CT images) to a low-dimensional feature space and, more importantly, its reverse (see Figure \ref{fig:GIN}). 
In contrast, GAN only finds the one-direction mapping from the feature space to the image space (i.e. generating), which makes it difficult to build the connection between the input images and the physical meaning of the feature space. 
In Section 3, we performed a case study using GIN to find the bidirectional feature mapping for the patients who underwent TAVR with the pre-surgical CT images as the input. 
Using the reverse mapping CNN, important clinical markers for the prediction of TAVR outcomes, such as the annular calcification, have been captured by the low-dimensional feature space (see Figure \ref{fig:result1}). Moreover, a test of predicting the surgical outcome directly using the selected features results in a high accuracy (see Figure \ref{fig:Pred}). This shows GIN preserves the pathophysiologically meaningful features while conducting the dimension reduction and can generate virtual patients with different possible surgical outcomes.

\vspace{-10pt}

\section{Methodology}

\vspace{-5pt}
\subsection{Preparing TAVR dataset with augmentation}
\vspace{-5pt}

\begin{figure}[!t]
\vspace{-35pt}
\centering
\includegraphics[width=0.75\textwidth]{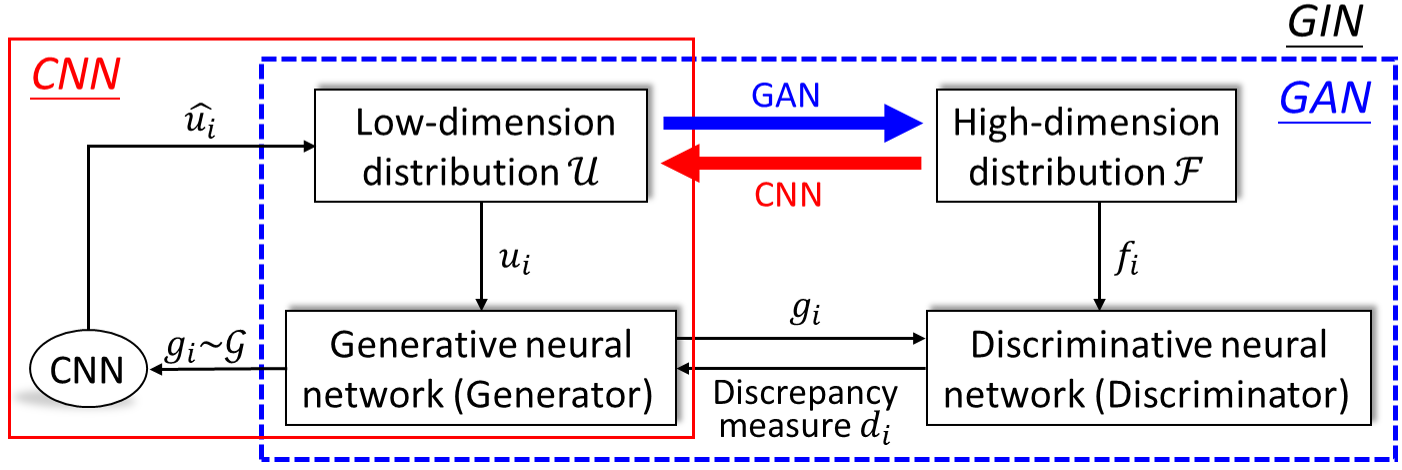}
\vspace{-10pt}
\caption{\label{fig:GIN} The overall architecture of GIN. It contains a GAN and a CNN.}
\vspace{-18pt}
\end{figure}
Aortic stenosis (AS) is one of the most common yet severe valvular heart diseases. Transcatheter aortic valve replacement (TAVR) is a less-invasive treatment option for AS patients who have a high risk of open-heart surgery \cite{conti2010biomechanical}. The deployment of the TAVR prosthesis involves a complex interaction between the prosthesis, the native aortic root, and the blood flow, which are not fully understood and may affect the procedural outcome, such as the degree of paravalvular leakage (PVL) and the risk of thrombosis/stroke \cite{conti2010biomechanical}. We studied the pre-procedural CT images of 168 AS patient (with an average age of 78) who received TAVR using a self-expandable prosthetic valve (CoreValve, Medtronic) from 2013 to 2016. All of the patients had pre-TAVR contrast-enhanced CT scans, which were performed on a 320-detector row CT scanner (Aquilion ONE, Toshiba). CT images were reconstructed with 10\% increments throughout the cardiac cycle, and the cardiac phase of the peak aortic valve opening was used. Each CT dataset contained a 3D volume of the cardiothoracic region. For computational purpose, we chose only one slide at the aortic annulus (selected by a
clinician) for this study. The method itself can be easily generalized to the 3D image volume. 
Post-TAVR PVL was set to be the major endpoint and 
was dichotomized to two groups: group 1 included none or low (trace to mild) PVL, while group 2 included high (moderate to severe) PVL.

We preformed routine data augmentation by slightly rotating the annular plane to add more samples. The regions of interest were rotated in 3D by four rotation angles in the annulus plane and one rotation angle in the longitudinal X-Z plane, from the original orientation. This led to an augmentation of 10 times the training set size. The augmented dataset was used to train the GIN.

\vspace{-10pt}
\subsection{Starting from GAN}
\vspace{-5pt}

The architecture of the GAN is shown in the blue dash box of Figure \ref{fig:GIN} \cite{goodfellow2016deep}. The key idea of the image generation by GAN is regarding the training set images as realizations of a distribution $\mathcal{F}$, which has extremely high-dimensional support (i.e. number of pixels of images). The distribution $\mathcal{F}$ can be physically interpreted as the group of images we are interested in (e.g., the aortic annulus). GAN can actually find a transformation from an easy-to-generate distribution $\mathcal{U}$ (usually, multi-uniform) to a distribution $\mathcal{G}$, which eventually is close enough to the target $\mathcal{F}$. In particular, GAN contains 2 neural networks (NN, see blue dash box of Figure \ref{fig:GIN}). 
In each training step of stochastic gradient descent (SGD), the realizations $u_i$ of $\mathcal{U}$ is fed into the \textit{generator} to generate $g_i$ following $\mathcal{G}^{(i)}$. Generated image $g_i$ is fed into the \textit{discriminator} to be compared with the training set data $f_i$ and find the discrepancy $d_i$, which is served as the loss function for the generator. The two NN's are trained by alternative optimization, until we think the generated distribution $\mathcal{G}$ is close enough to the true distribution $\mathcal{F}$. 

GIN contains a GAN part for generation (blue dash box of Figure \ref{fig:GIN}). Moreover, in our framework, the support of distribution $\mathcal{U}$ is regarded as the feature space (it does not yet have any physical meaning) and realizations of the distribution $\mathcal{U}$ are the hidden features of the corresponding valves. This means given a feature vector (a realization of distribution $\mathcal{U}$), the GAN part in GIN can generate a virtual valve based on that feature vector.

\vspace{-10pt}
\subsection{Adding a CNN for reverse mapping}
\vspace{-5pt}

As mentioned, the generation using only GAN lacks pathophysiologic interpretation. The reason is that it only gives one-direction mapping from the feature space $\mathcal{U}$ to the real valve distribution $\mathcal{F}$ (assuming the final $\mathcal{G}$ is close enough to the true distribution $\mathcal{F}$, see Section 2.2). Thus, the feature space itself is difficult to interpret, and we are generating virtual patients without meaningful guidance. One way to introduce the pathophysiologic meaning to  the feature space is to find corresponding locations of the real patients in that space, since the real patients have surgical records, such as the post-TAVR PVL level, which can be used to label the space and conduct classification. In other words, we need to find the backward mapping from the real valve distribution $\mathcal{F}$  to the feature space $\mathcal{U}$. Therefore, besides GAN, we add a CNN to the framework regarding the generated images $g_i$ from $\mathcal{G}$ as input and the feature $u_i \sim\mathcal{U}$ as the output (see red box of Figure \ref{fig:GIN}). After the CNN is trained, we may feed the model with real patients data $f_i \sim \mathcal{F}$ and find its corresponding feature in the feature space.

In most literature, CNN is used for classification \cite{shin2016deep}, which means the supervised value for each data set is discrete. Here, we use the CNN for regression, which means the label $u_i \sim \mathcal{U}$ (features) is a continuous vector with a non-zero measure. This is much more difficult for training when the dimension of $\mathcal{U}$ is high. But the advantage is that we are using the realization of distribution $\mathcal{G}$ (instead of $\mathcal{F}$) as the training set, in which, theoretically speaking, the available data size is infinitely large. In reality, we restrict the dimension of $\mathcal{U}$ to be less than 20 (10 in the case study) to gain a stable training result from CNN.

\vspace{-10pt}
\subsection{GIN framework}
\vspace{-5pt}

Putting everything together, GIN contains three NN's, two of them first form a GAN (one generator and one discriminator) to find the transformation from the feature space to the CT images space, then the other NN finds the reverse mapping from the CT images space to the feature space (see Figure \ref{fig:GIN}). Finally, we have the bidirectional mapping between features and CT images. Furthermore, the feature space selected by GIN captures the pathophysiologic information hidden in CT images, which can be used to predict surgical complications (PVL). This allows us to conduct arithmetic operations in the feature space and make sure any generated virtual patients have physical and pathophysiologic meanings (e.g., we can generate a virtual patient knowing it may lead to high PVL or not). 

It is important to note that our method is essentially different from adversarially learned inference (ALI) or  bidirectional GAN (BiGAN) \cite{donahue2016adversarial}  in the literature. In order to invest the feature space with pathophysiologic meanings, we need a \textit{hard} inverse, i.e. CNN=Generator$^{-1}$ for every input sample. Thus, GIN has a sequential order of GAN and CNN to make sure the sample-to-sample inverse is explicitly trained and thus has better expressibility (see reconstruction test in Section 3.1). In contrast, BiGAN or ALI uses one discriminator to supervise both generator and encoder, the generator and encoder would be inverse to each other, as claimed, yet only in distribution level, which is not rigorous enough for medical image applications. Moreover, it uses coupling training of 3 NNs. This complicated architecture requires more fine tuning and therefore less suitable for our sparse dataset (see Section 2.1).

\begin{figure}[!t]
\vspace{-20pt}
\centering
\includegraphics[width=0.75\textwidth]{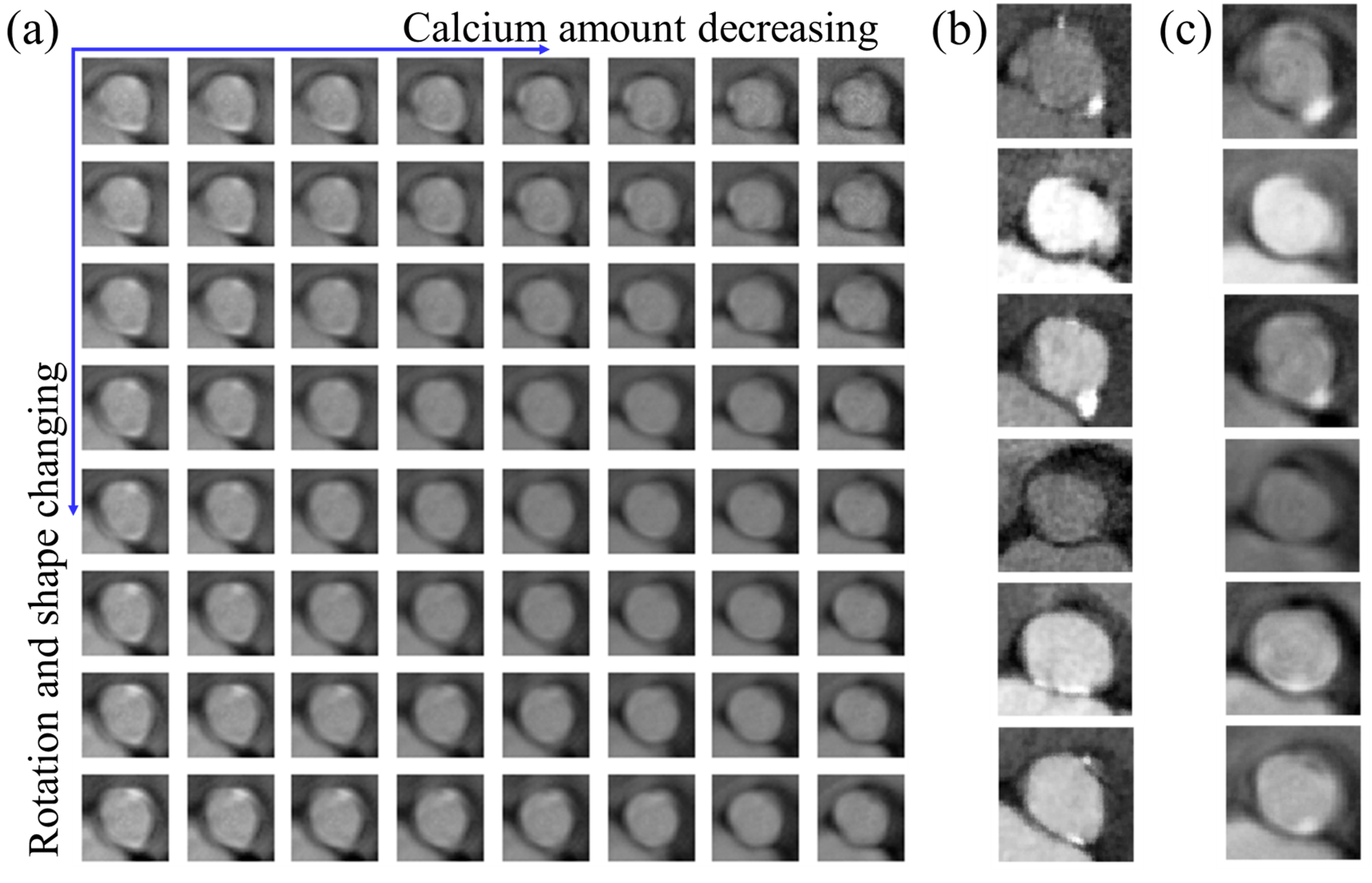}
\vspace{-10pt}
\caption{\label{fig:result1} The training results of GIN. (a) the characteristic valve CT images in the feature space, (b) The real patients valve CT image, (c) reconstruction test of the real valves in (b).}
\vspace{-20pt}
\end{figure}

\vspace{-15pt}
\section{Results}

\vspace{-5pt}
In this test, the dimension of the feature space $\mathcal{U}$ is chosen to be 10, the results can be sharper if the feature dimension is increased to 20. But the training cost will also increase dramatically. The two NN's of the GAN part adapt 2-layer vanilla neural networks with 512 hidden nodes in each hidden layer and ReLu activations. CNN has approximately the same complexity with leaky ReLu activation and batch normalization in each layer (see \cite{arjovsky2017wasserstein} for more details).

\vspace{-10pt}
\subsection{Pathophysiology-interpretable feature mapping}
\vspace{-5pt}
After training the GIN, a 2D cross-section in the feature space of the valves are shown in Figure \ref{fig:result1} (a). The small figures at different locations mean the corresponding characteristic valve CT images in the specific locations of the feature space. We may find some physical meaning for the two features. In every column, from top to bottom, the valve rotates clockwise and the shape of the valve wall is gradually changed. In every row, from left to right, the amount of calcification (which is the brightest region in the CT images) decreases. According to clinical observations, high amounts of annular calcification could be an important risk factor of
post-TAVR 
PVL. Thus, we may speculate that the left region in the feature space, which has visually more calcium, may be associated with higher rates of surgical complications.

Since the bidirectional mapping between the feature space and the valve space (see Section 2.4) is found by GIN, We may conduct the following reconstruction test to visualize the information loss by the framework. The features of the real patients' CT images were first extracted by the CNN part, and then the extracted features were used to generate virtual CT images by the GAN part. Ideally, if there is no information loss in both feature extraction (CNN) and generation (GAN), the reconstructed images should be identical to the real ones. The test results of some representative real CT images (Figure \ref{fig:result1} (b)) are shown in Figure \ref{fig:result1} (c). In the test, the reconstructed images look similar to their real counterparts, especially the overall shape and orientation of the valve. Meanwhile, some of the important details like calcification are also captured. This shows that the GIN captures pathophysiologically meaningful features. 
Yet some of the details are missing and also the reconstructed images are not as sharp as the real ones. This may be because the training set data is too small even with the augmentation to generate high fidelity images and the feature space is set to be too low to capture higher order features. Comparing our reconstruction test and the ones in the BiGAN paper \cite{donahue2016adversarial}, we would conclude that GIN is better in extracting the features and finding a sample-to-sample hard inverse.

\begin{wrapfigure}{r}{0.35\textwidth} 
\vspace{-50pt}
  \begin{center}
    \includegraphics[width=0.35\textwidth]{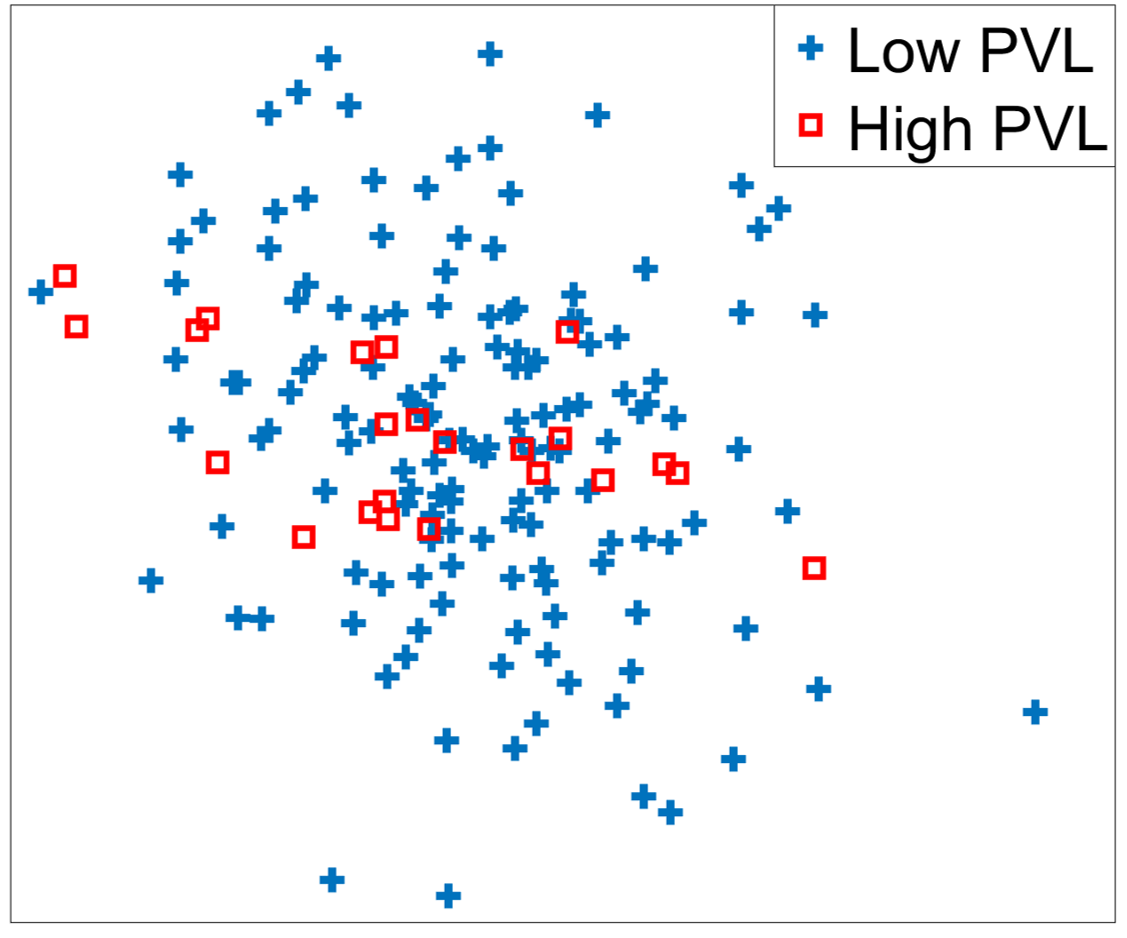}
    \vspace{-20pt}
    \caption{The feature mapping of the real patients in the feature space with different PVL levels. }
    \label{fig:ISO}
  \end{center}
  \vspace{-30pt}
  \vspace{1pt}
\end{wrapfigure} 


\vspace{-10pt}
\subsection{Post-TAVR PVL prediction}
\vspace{-5pt}

In order to assess the pathophysiologic meaning of the feature space, we look for the relationship between the selected features and PVL. The first 2 Isomap \cite{tenenbaum2000global} features are shown in Figure \ref{fig:ISO}, where the red squares represent the patients with high PVL and the blue crosses represent the patients with low PVL. 
The two groups of different PVL levels follow different, visually distinguishable distribution, even projecting to a 2D feature plane.
\begin{wrapfigure}{r}{0.36\textwidth} 
\vspace{-18pt}
  \begin{center}
    \includegraphics[width=0.36\textwidth]{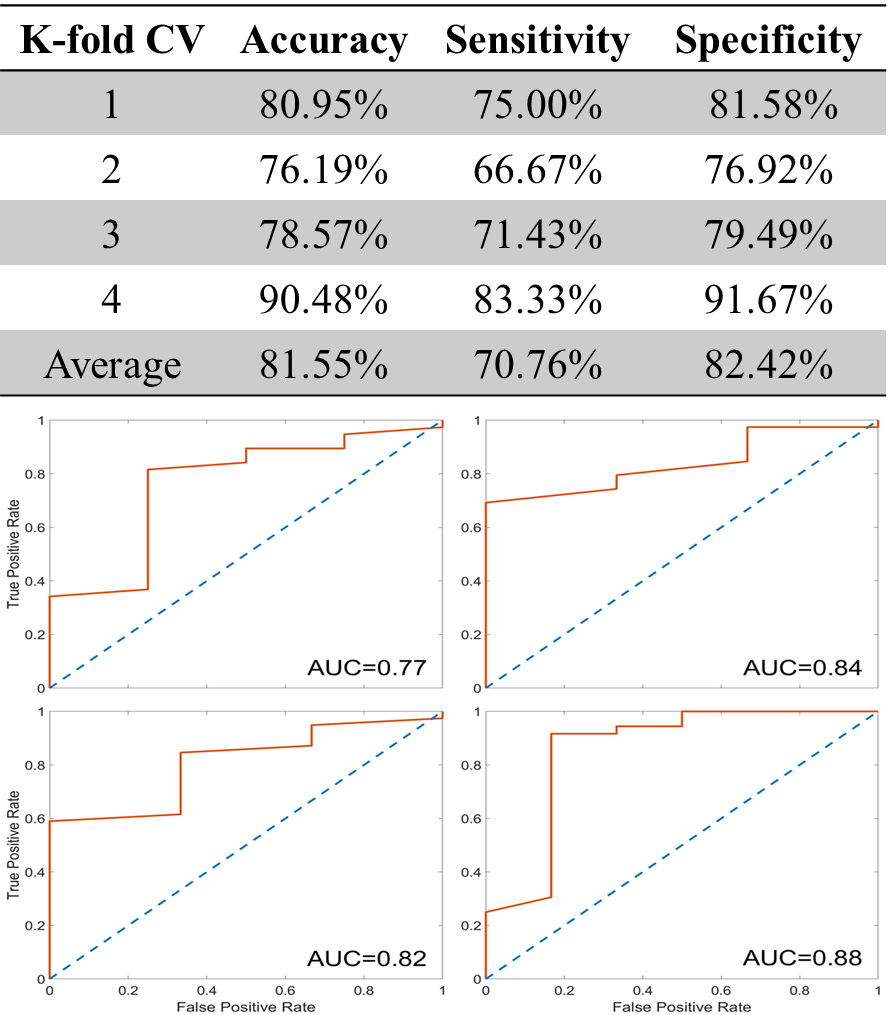}
    \vspace{-20pt}
    \caption{The accuracy measurements (upper) and ROC curves (bottom) of the random forest model in predicting PVL.}
    \label{fig:Pred}
  \end{center}
  \vspace{-35pt}
\end{wrapfigure} 

A more rigorous approach is to quantify the pathophysiologic significance by predicting the post-TAVR PVL level using the features selected. 
A simple random forest classifier (total 500 decision trees) was used to classify the two groups, namely high PVL and low PVL. A 4-fold cross validation (75\% of data as a training set and 25\% as a validation set) was adopted to check the prediction performance as shown in Figure \ref{fig:Pred}. The average of the test accuracy, sensitivity, and specificity were 81.55\%, 70.76\%, and 82.42\% respectively. The receiver-operating characteristic (ROC) curves are shown in Figure \ref{fig:Pred} of each validation and the AUC values are 0.77, 0.84, 0.82, and 0.88 respectively. All of these turned out to be statistically significant ($p<0.001$). This promising result shows that the features selected by GIN is pathophysiologically interpretable and the information related to PVL outcomes in CT image is well-preserved. 

\vspace{-10pt}
\subsection{CT image generation}
\vspace{-5pt}

\begin{wrapfigure}{r}{0.3\textwidth} 
\vspace{-50pt}
  \begin{center}
    \includegraphics[width=0.3\textwidth]{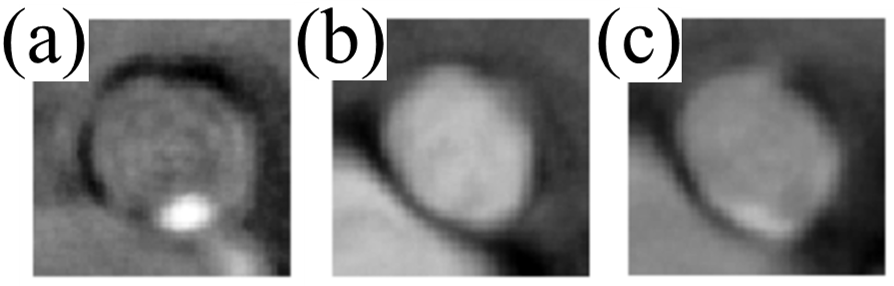}
    \vspace{-20pt}
    \caption{Virtual patient generation with possibly different PVL levels.}
    \label{fig:gener}
  \end{center}
  \vspace{-30pt}
  \vspace{1pt}
\end{wrapfigure}


More importantly, the pathophysiologically interpretable features captured by GIN can be used for virtual patient generation. Recall that the GAN can only generate the virtual patients that look like real patients. However, GIN can generate virtual patients with specific pathophysiologic appearances. The random forests classifier (see Section 3.2) actually segments the feature space to two parts according to its predicted PVL level. Thus, we may generate a virtual patient with a high probability of resulting in a high PVL by selecting a feature vector in the high PVL part of the space. As shown in Figure \ref{fig:gener} (a), the generated CT image visually contains a large calcified nodule, which may lead to a high level post-TAVR PVL. We can also generate a virtual patient that is most likely with a low or none PVL as shown in Figure \ref{fig:gener} (b). Also, we may generate a virtual patient with the features near the decision boundary as shown in Figure \ref{fig:gener} (c). 
Despite the high prediction accuracy shown in Figure \ref{fig:Pred}, the sensitivity is relatively low. Thus, we may generate more virtual patients with a high PVL (Figure \ref{fig:gener} (a)) to reduce the imbalance outcome of the dataset. Also, generating virtual patients with the features near decision boundary (Figure \ref{fig:gener} (c)) can be extremely helpful to improve the prediction ability of the future predictive model. The generated virtual patients can then be 3D printed and go through virtual surgeries to obtain the PVL label in vitro (see \cite{qian2017quantitative} for more experimental details) as future work.



\vspace{-12pt}
\section{Conclusion}
\vspace{-5pt}

We proposed a new generative framework \texttt{-} GIN \texttt{-} to generate visually authentic virtual patients by finding the bidirectional feature mapping between the features and the real CT images (see Figure \ref{fig:result1}). Moreover,
a test of predicting the surgical outcome directly using the selected features resulted in a high accuracy, which suggests that features contain pathophysiologic meaning
(see Figure \ref{fig:Pred}).
This means GIN can generate virtual patients with different surgical outcomes for later 3D printing and in-vitro experiments (see Figure \ref{fig:gener}). 
These virtual patients can be crucial in enhancing the model prediction power as an additional data source and more importantly, understanding the nature of the disease and performing optimal pre-surgical planning.
In general, applying GIN to generate physically interpretable virtual samples has great potential for image related machine learning methods with limited and unbalanced datasets.

%



\vspace{-12pt}

\end{document}